\documentclass[a4paper,fleqn]{cas-dc}

\usepackage[numbers]{natbib}
\usepackage{algorithm}
\usepackage{algpseudocode}
\usepackage{amsmath}
\usepackage{graphics}
\usepackage{epsfig}
\usepackage{url}
\def\tsc#1{\csdef{#1}{\textsc{\lowercase{#1}}\xspace}}
\tsc{WGM}
\tsc{QE}
\tsc{EP}
\tsc{PMS}
\tsc{BEC}
\tsc{DE}

\begin{document}
	\let\WriteBookmarks\relax
	\def\floatpagepagefraction{1}
	\def\textpagefraction{.001}

	\shortauthors{Zhe Chu et~al.}
	\title [mode = title]{Robotic grasp detection using a novel two-stage approach} 
	
	\tnotemark[1]   
	\tnotetext[1]{We gratefully acknowledge the assistance of Haobin Shi and Guosheng Huang giving us some guidance.}               
	\author[1]{Zhe Chu}
	
	\credit{Main undertaker of the research.Responsible for completing experiments and writing papers}
	
	\address[1]{Northwestern Polytechnical University,China}
	
	\author[2]{Mengkai Hu}

	\credit{Conceptual design and data collection}
	
	\address[2]{Peking University, China}
	
	\author[3]
	{Xiangyu Chen\corref{mycorrespondingauthor}}
	\credit{Make important changes to the paper and approve the final paper to be published}
	\cortext[mycorrespondingauthor]{Corresponding author}
	\ead{chxy95@mail.nwpu.edu.cn}
	\address[3]{Northwestern Polytechnical University,China}

\begin{abstract}
Recently, deep learning has been successfully applied to robotic grasp detection. Based on convolutional neural networks (CNNs), there have been lots of end-to-end detection approaches. But end-to-end approaches have strict requirements for the dataset used for training the neural network models and it’s hard to achieve in practical use. Therefore, we proposed a two-stage approach using particle swarm optimizer (PSO) candidate estimator and CNN to detect the most likely grasp. Our approach achieved an accuracy of 92.8\% on the Cornell Grasp Dataset, which leaped into the front ranks of the existing approaches and is able to run at real-time speeds. After a small change of the approach, we can predict multiple grasps per object in the meantime so that an object can be grasped in a variety of ways.
\end{abstract}

\begin{keywords}
Robotic grasp detection \sep Convolutional neural network \sep Two-stage cascaded system \sep Particle swarm optimizer
\end{keywords}

\maketitle

\section{Introduction}

Robotic grasping is a core function for intelligent robot to perform a variety of autonomous manipulation tasks\cite{1}. Humans can instinctively perceive the unstructured environment, find out the characteristics of the grasped object and grasp the object directly. But the robot can’t do that and if we model the surrounding environment around the robot, it is not only time-consuming but also hard to model.

Recently, with the development of deep learning,many end-to-end robotic grasp detection approaches based on CNN are developed. Most of these new approaches are one-stage approaches, which find a good grasp in one step using CNN and get a high accuracy on the datasets. But most of these end-to-end approaches\cite{7820198}\cite{8202237}\cite{2} get the results in one step using trained CNN models, so those approaches have very high request to accuracy of the model which relies heavily on the quality of datasets. But in practical use, it’s very hard to make a high-quality dataset. Hence, in order to garner better results in practice, we need to find a robotic grasp detection approach which has a lower request to accuracy of the model.

Sliding window detection is a robotic grasp detection which was often used in the past. The approach uses the classifier to test the selected parts of the image one by one. The part which has the highest score is considered to be the best grasp of the object\cite{2}. Although this approach is a little time-consuming, it can take measures such as randomization, conditional constraints and so on to reduce the influence of model’s inaccuracy on the final result.

Based on sliding window detection and CNN, we propose a two-stage robotic grasp detection approach which can extract the best grasp from the object's RGB image using PSO candidate estimator and CNN. First, we design a robotic grasp identification model based deep CNN to determine whether the input is a good grasp. Then, we find good candidate grasps using candidate estimator. This candidate estimator finds proper candidate grasps using PSO algorithm. Using this approach, we lower the requirement of the model accuracy. We evaluate our approach on the Cornell Grasp Detection Dataset (see Figure 1). Our approach achieved an accuracy of 92.8\%. 

This paper is organized as follows: in section 2, we discuss the related work. In section 3, we present our identification models. In section 4, we describe our candidate estimator. In section 5, we present our experiment and evaluation. In section 6, We present our results and then compare it with other approaches. In section 7, we present our experiment on real robots. Finally, we provide our conclusions.
\begin{figure}[h]
	\centering
	\includegraphics[scale=.75,width=0.4\paperwidth]{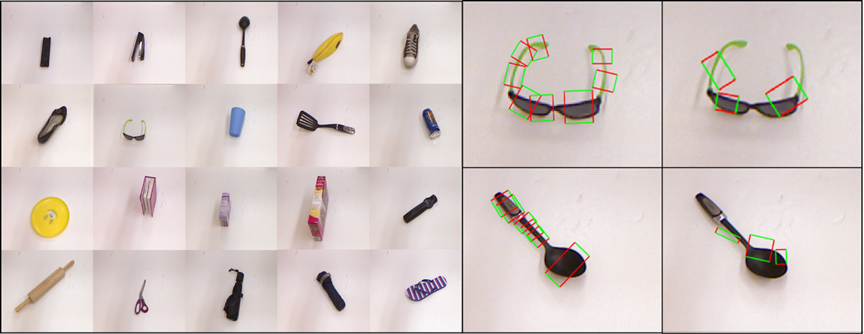}
	\caption{Cornell Grasp Detection Dataset contains a variety of grasp objects, every object has many graspable feature labels and ungraspable feature labels.}
	\label{FIG:1}
\end{figure}

\section{Related work}
In the past few decades, with the development of robots and artificial intelligence, a number of robotic grasp detection approaches have been produced. Most of the early approaches depend on the accurate information about environment and object to find good grasps. Based on accurate information, some approaches\cite{3}\cite{1371616}\cite{4}\cite{1308797} are used to realize successful and stable grasps. But the disadvantage of these approaches is that many robots usually don’t know beforehand the model of the grasp object, so it’s hard to build the complex 3D models of the objects.

Recent years, with the development of deep learning, many researchers propose to detect the grasp using deep learning. Here are the known approaches.

Ian Lenz proposed a two-stage approach based on sliding window detection\cite{1}. First, they find some candidate grasp rectangles. Then they test all these candidate grasp rectangles using a CNN with simple structure to get a set of grasps with high probability. Finally, they use a complex CNN to rank all the grasps which are screened out for the first stage and find the best grasp of the object. Because the sliding window detection is too time-consuming, Zhichao Wang proposed a novel robotic grasp detection system\cite{doi:10.1177/1687814016668077}. In this system, they first reuse the result of the former grasping detection for feedback to find candidate grasp with higher probability. Then they use a CNN to evaluate the candidate grasps and find the best grasp.

Joseph Redmon proposed a new CNN model by simplifying AlexNet network\cite{2}. Using this model, they can directly get a regression to the grasp and the classification of the grasp object. With the development of deep learning, Sulabh Kumra proposed a novel CNN model for robotic grasp detection\cite{8202237}. This model consists of two 50-layer CNNs and a shallow CNN. They merge the outputs of these two 50-layer CNNs and input them into the shallow CNN to predict the grasp configuration.

Among these approaches, Ian Lenz and Zhichao Wang’s approaches are two-stage, but the accuracy is not good (see Table 1). Joseph Redmon and Sulabh Kumra’s approaches are end-to-end,which have high requirements for the accuracy of the dataset. It’s hard to achieve in practical use. So we decide to attempt a better two-stage approach.
\section{Identification model}

We use CNN as the grasp identification model. In order to enable the identification model to learn how to distinguish between grasping and non-grasping features, we first train the network as a binary classifier. After the model training, the classification probability of the softmax layer is taken as the model output. In order to reduce the computing cost and increase the computing speed, we used smaller networks and tried the following models.
\subsection{Simplified AlexNet\cite{NIPS2012_4824} model}

Our first model is based on a simplified version of Alex\-Net proposed by Krizhevsky. AlexNet achieved high accuracy on target recognition tasks. Our first model used the AlexNet architecture, but changed the configuration. The model (see Figure 2) includes 3 convolutional layers,3 maximum pooling layers, and 2 fully connected layers. Finally, softmax activation function is used to activate the output. 
\begin{figure}
	\centering
	\includegraphics[scale=.75,height=0.18\paperwidth,width=0.4\paperwidth]{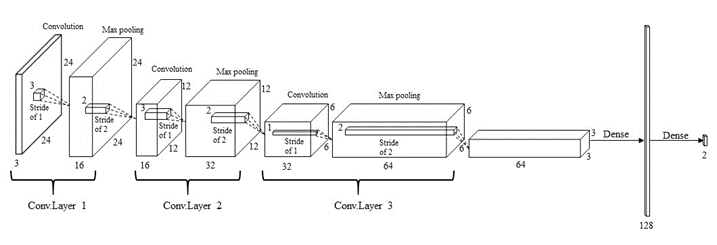}
	\caption{Simplified AlexNet model.}
	\label{FIG:2}
\end{figure}

We trained the model. And the highest training accuracy of the model is 83.39\%, and the highest validation accuracy is 82.12\%, which is not ideal. We use the method of feature map visualization to find the problems existing in the model. We output the feature map behind each layer (see Figure 3). From the input image on the far left of figure 3, we can see that the key features of the image should be located in the middle of the image. But in the end the critical feature our model learned (the yellow part of the image) was to the left of the image.
\begin{figure}[h]
	\centering
	\includegraphics[scale=.75,width=0.4\paperwidth]{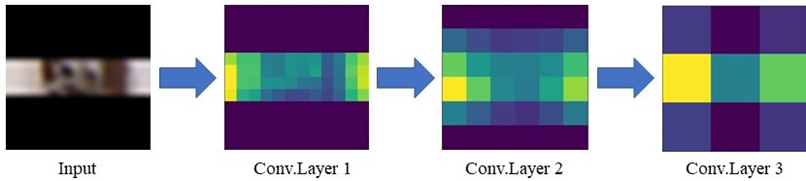}
	\caption{The result of Simplified AlexNet model’s feature map visualization.}
	\label{FIG:3}
\end{figure}

This shows that our model only extracts part of the texture details, while the critical features of the image are not extracted. In addition, due to the existence of pooling layer, the further back, the less image information will be retained.
\subsection{Original GraspingNet model}

Based on the problems with Simplified Alexnet model, we decided to try using a more complex model to extract deep, critical features. We increase the size and number of convolution kernels. In the new model (see Figure 4), the size of the convolution kernel of the three convolutional layers is 3, 5 and 7, and the number is 32,64,128, respectively. In addition, we remove the Maxpooling layer after each convolutional layer to retain more information. And we add the Batch Normalization operation between the convolution and activation functions of the convolution layer to speed up convergence. After the two fully connected layers, we use the softmax activation function to output. Finally, we select Relu as the activation function of the convolution layer.
\begin{figure}[h]
	\centering
	\includegraphics[scale=.75,height=0.18\paperwidth,width=0.4\paperwidth]{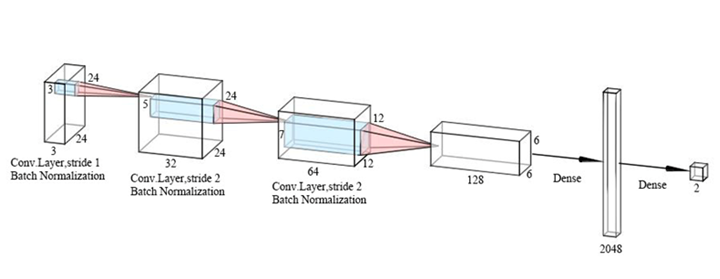}
	\caption{Original GraspingNet model.}
	\label{FIG:4}
\end{figure}

We trained the model and achieved 98.50\% train accuracy and 85.50\% validation accuracy. It is obvious that the model has the phenomenon of overfitting. And this modelis time-consuming, it takes four times as long as the first model. In addition, during the process of feature map visualization (see Figure 5), we found that Original GraspingNet model was still not so ideal in extracting critical features. As can be seen from Figure 5, the critical features extracted by the model are still to the left of the picture.
\begin{figure}[h]
	\centering
	\includegraphics[scale=.75,width=0.4\paperwidth]{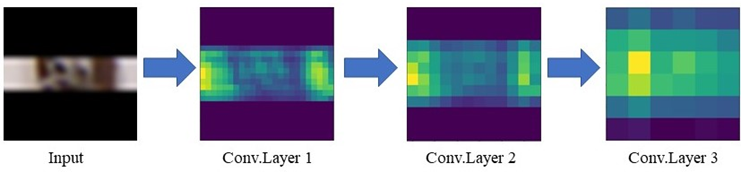}
	\caption{The result of Original GraspingNet model’s feature map visualization.}
	\label{FIG:5}
\end{figure}
\subsection{Final GraspingNet model}
We modified our identification model based on the former 2 models. 
A CNN usually consists of convolution layer, pooling layer and full connection layer. But in Final GraspingNet model, we also remove the pooling layer and add the Batch Normalization operation between the convolution and activation functions of the convolution layer. Then, we remove the full connection layer with a large number of participants and use the GAP layer instead. By doing so, on the one hand, it can prevent overfitting and make the classification more natural; on the other hand, it can also increase the speed of the model. So our final identification model is an 8-layer CNN, which is mainly composed of Convolution layer, Global Average Pooling (GAP) layer and the output layer (see Figure 6).
\begin{figure}
	\centering
	\includegraphics[scale=.75,height=0.15\paperwidth,width=0.4\paperwidth]{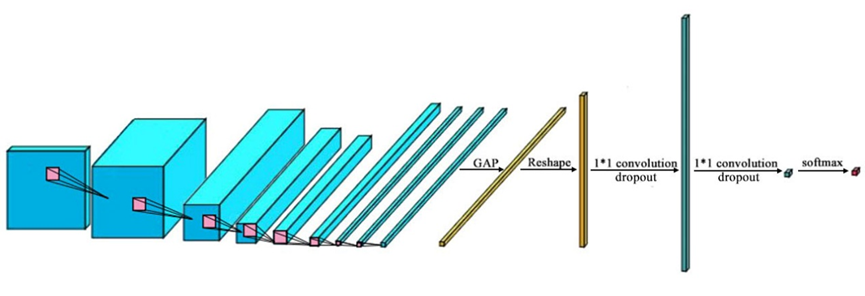}
	\caption{ The architecture of the identification model. There are 8 convolution layers in the hidden layer. And in each convolution layer, we do convolution operation, Batch Normalization operation and Relu operation.}
	\label{FIG:6}
\end{figure}
\subsubsection{Convolution layer}
In this model, we decide to replace the large convolution kernels of Original GraspingNet with 3 by 3 convolution kernels so that it can improve the ability of the model to extract critical features while increasing the speed of the model. In each convolution layer, the model needs to perform convolution, Batch Normalization, and Relu activation function operations in turn, before being input to the next convolution layer. 

In the first convolution layer, we use 32 3*3 convolutional kernels to calculate with stride 1, and ignore the pixels of the image edge by setting padding=valid. After the convolution operation, we'll do the Batch Normalization and Relu activation function operations. The operation of the first convolution layer is defined as
\begin{math}
F_{c1}(.),
\end{math}
the input is defined as 
\begin{math}
I_{c1},
\end{math}
and the output is defined as 
\begin{math}
R_{c1}.
\end{math}
Then the operation of the first convolution layer is as follows:
\begin{flalign}
&&R_{c1}=&F_{c1}(I_{c1})&
\end{flalign}

In the second, third and fourth convolution layers, we use 64 3*3 convolution kernels with stride 2 for operation. After the convolution operation, we'll do the Batch Normalization and Relu activation function operations. This operation is defined as
\begin{math}
F_{c2}(.), 
\end{math}
and the output of the second, third and fourth layer is defined as
\begin{math}
R_{c2},R_{c3},R_{c4},
\end{math}  
then the operation of the second, third and fourth convolution layer is as follows:
\begin{flalign}
&&R_{c2}=&F_{c2}(R_{c1})& \\
&&R_{c3}=&F_{c2}(R_{c2})& \\
&&R_{c4}=&F_{c2}(R_{c3})& 
\end{flalign}

In the last four convolution layers, we use 128 3*3 convolution kernels to operate with stride 2. This operation is defined as 
\begin{math}
F_{c3}(.),
\end{math} 
and the output of the last four convolution layers is defined as
\begin{math}
R_{c5},R_{c6},R_{c7},R_{c8},
\end{math}
 then the operation of the last four convolution layers is as follows:
\begin{flalign}
&&R_{c5}=&F_{c3}(R_{c4})& \\
&&R_{c6}=&F_{c3}(R_{c5})& \\
&&R_{c7}=&F_{c3}(R_{c6})& \\
&&R_{c8}=&F_{c3}(R_{c7})& 
\end{flalign}

\subsubsection{GAP layer}
The Global Average Pooling operation conducted by GAP layer is defined as
\begin{math}
F_{GAP}(.),
\end{math}
and the output is defined as
\begin{math}
R_{GAP},
\end{math}
then,
\begin{flalign}
&&R_{GAP}=&F_{GAP}(R_{c8})&
\end{flalign}
\subsubsection{Output layer}
Before the final output of the model, we first use the convolution of 1*1 to perform ascending and reducing operations on the output of the GAP layer to increase the information combination across channels. In addition, we will use dropout to further prevent overfitting of the model before ascending and reducing operations. 

We define the dropout operation as
\begin{math}
F_{D}(.),
\end{math}
the 1*1 convolution ascending dimension operation as
\begin{math}
F_{A}(.)
\end{math}
and the output of the dropout operation and ascending dimension operation as
\begin{math}
R_{A},
\end{math}
then,
\begin{flalign}
&&\quad   R_{A}=&F_{D}(F_{A}(R_{GAP}))&
\end{flalign}

We define 1*1 convolution reducing operation as
\begin{math}
F_{R}(.)
\end{math}
and the output of the dropout operation and ascending dimension operation as
\begin{math}
R_{R},
\end{math}
then,
\begin{flalign}
&&\quad \,  R_{R}=&F_{D}(F_{R}(R_{A}))&
\end{flalign}

Finally, we activate it using the softmax function, which outputs the results of the model identification in probabilistic form. We define the operation of the softmax function as
\begin{math}
F_{S}(.)
\end{math}
and the output of the output layer as
\begin{math}
O,
\end{math}
then,
\begin{flalign}
&&\quad \  O=&F_{S}(R_{R})&
\end{flalign}

\subsubsection{Model training}
The training process of the convolutional neural network model is to find the global optimal solution in the parameter space. In this process, we may encounter many local optima, but we have to skip the local optima to find the global optima in the parameter space to train the best model.

Therefore, we used the stochastic gradient descent me\-thod to train the CNN model and set the learning rate in sections. During training (see Figure 7), the learning rate decayed every 60 epochs. In this way, when the model starts training, because the training speed is faster, some local optima are skipped. Finally, when the model is trained near the optimal solution, the learning rate converges to the optimal solution of the model at a very low learning rate. In addition, during the training, we also increased the value of momentum. On the one hand, it is more helpful for the model to get rid of the local optimal solution, on the other hand, it can also increase the stability of model training to some extent.
\begin{figure}[h]
	\centering
	\includegraphics[scale=.75,width=0.4\paperwidth]{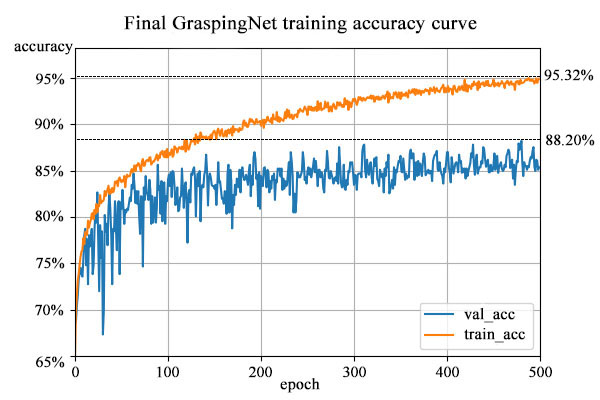}
	\caption{Final GraspingNet training curve}
	\label{FIG:7}
\end{figure}
\section{PSO candidate estimator}
\subsection{Five-dimensional representation}
There are mainly two kinds of grasp representation. One is a seven-dimensional representation which is proposed by Jiang Yun\cite{5980145}. The other is a five-dimensional representation which is proposed by Ian Lenz\cite{1}. The five-dimensional representation is a simplification of the seven-dimensional representation. In order for computational convenience, we use the five-dimensional representation. In the five-dimen\-sional representation, the grasp is a rectangle which is determined by location, size and direction as:
\begin{center}
\begin{math}
g = \left\{ x,y,\theta,h,w \right\}
\end{math}
\end{center}
\begin{math}
(x,y)
\end{math}
 is the location of a rectangle,
\begin{math}
\theta
\end{math}
is the direction of the rectangle relative to the horizontal axis,h is the length of the gripper,w is the width of the gripper opening(see Figure 8).
\begin{figure}[h]
	\centering
	\includegraphics[scale=.75,width=0.4\paperwidth]{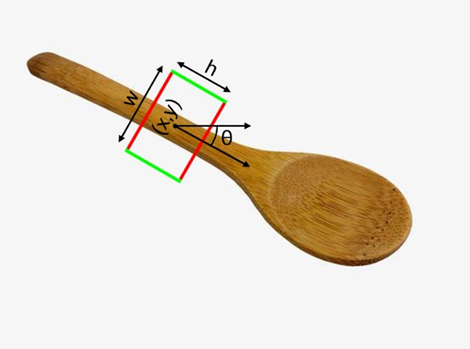}
	\caption{five-dimensional grasp representation}
	\label{FIG:8}
\end{figure}
\subsection{Candidate estimator}
In the sliding window detection, every image to be detected is a large and continuous state space. If we use enumeration approach to search the grasp, we need to detect a great number of candidate grasps. For example, if we detect an image whose size is 224 by 224, we may need to consider 50176(224*224=50176) locations. Suppose that there are 70 different lengths of grippers, the width of the gripper opening ranges from 30 to 100 and that the angle of the gripper ranges from 0° degree to 180° and the angle changes in one degree, we should consider 632,217,600(50176*70*180 = 632,217,600) candidates. If we do that, the cost of computing is very large.

In order to solve this problem, we design a candidate estimator based on PSO. This estimator transfers the search of the grasp to an optimization problem. The parameters of the optimization problem are the five-dimensional representation of every candidate. The processed image is the state space of the optimization problem and the constraint of the problem is the edges of the image. Our optimization goal is to find a candidate to optimize the score function of the grasp. As for the score function, we use the grasp identification model in section 3. As can be seen above, the grasp identification model outputs the probability of grasping. Then, we use PSO to solve the problem.

\subsection{PSO algorithm}
PSO is a kind of stochastic optimization technique based on population, which was proposed by Eberhart and Kennedy in 1995. PSO algorithm mimics the clustering behavior of animals such as insects, herds, birds, and so on. These populations search for food in a cooperative way. Every particle of the population constantly changes the way it searches by learning from its own experience and that of other members. During the search of the algorithm, each particle searches for the optimal solution separately in the search space, and marks it as the current individual extremum. Then, the particle shares its individual extremum with other particles in the particle swarm and the particle swarm finds the optimal individual extremum as the current global optimal solution. All particles of the particle swarm adjust their speed and position according to their current individual extremums and the current global optimal solution shared by the particle swarm. Therefore, the algorithm can be used to deal with the optimization problem of multivariable functions with multiple local optima.

In this article, we use a particle 
\begin{math}
x_{i}
\end{math}
to represent a candidate. Every particle has a speed
\begin{math}
v_{i}^{k}
\end{math}
at iteration 
\begin{math}
k.
\end{math}
The movement of the particle is determined by its current individual extremum
\begin{math}
p_{best_{i}}
\end{math}
and the current global optimal solution
\begin{math}
g_{best}.
\end{math}
The updated formula of particle velocity and position is as follows:
\begin{flalign}
&&v_{i}^{k+1}=wv_{i}^{k}+c_{1}&r_{1}p_{best_{i}}+c_{2}r_{2}g_{best}&\nonumber\\
&& x_{i}^{k+1}=&x_{i}^{k}+v_{i}^{k+1}& \nonumber
\end{flalign}
\begin{math}
w
\end{math}
is the inertia factor.
\begin{math}
c_{1}
\end{math}
and
\begin{math}
c_{2}
\end{math}
are accelerating factors.
\begin{math}
r_{1}
\end{math}
and
\begin{math}
r_{2}
\end{math}
are 2 random numbers, which are used for increasing the randomness of the search. The is shown in Algortithm 1.
\begin{algorithm}[h] 
	\caption{PSO Algortithm} 
	\begin{algorithmic}[1] 
		\State $g\_fit=0,initial\_time=0$.		
		\While {$(g\_fit<init)and(initial\_time<max\_init)$}
		 \State Randomly initialize particle swarm $x_{i}$,$v_{i}$;
		 \State Use identification model to caculate score for $x_{i}$,$v_{i}$;
		 \State Update $p\_fit_{i},g\_fit$; 
		 \State $initial\_time=initial\_time+1$;
	    \EndWhile
	    \While {$(g\_best<prob)and(iteration<max\_iter)$}
	    \State  $v_{i}^{k+1}=wv_{i}^{k}+c_{1}r_{1}p_{best_{i}}+c_{2}r_{2}g_{best}$;
	    \State $x_{i}^{k+1}=x_{i}^{k}+v_{i}^{k+1}$;
	    \State Use identification model to caculate score for $x_{i}$,$v_{i}$; 
	    \State Update $p\_fit_{i},g\_fit$;
	    \State Update $p\_best_{i},g\_best$;
	    \State $iteration=iteration+1$;
	    \EndWhile
		\State Output the best candidate grasp.   
	\end{algorithmic} 
\end{algorithm}

In Algorithm 1,$g\_fit$ is the highest score of all particles and $p\_fit$ is the highest score of every particle. $initial\_time$ is the iteration during initializing and $iteration$ is the number of searches. $max\_init$ is the maximum number of initializations and $max\_init$ is the maximum number of searches.$init$ is the minimum threshold that $g\_fit$ should meet during initialization and $prob$  is the minimum threshold that $g\_fit$ should meet during searching.Through this candidate estimator, we can find a good grasp fast. The iterative convergence process of particles is as follows (see Figure 9): the particles are randomly distributed inside the image during initialization. And as the algorithm iterates, the particles gradually gather near the optimal location. If the center of a particle is near the edges of the image during initialization or iteration so that some components of the particle exceed the boundaries of the image when it is segmented, the particle will be eliminated and we will add a new particle randomly. The convergence condition is that the number of iterations exceeds the maximum number of iterations or the global optimal particle adaptation value reaches the threshold.
\begin{figure}[h]
	\centering
	\includegraphics[scale=.75,width=0.4\paperwidth]{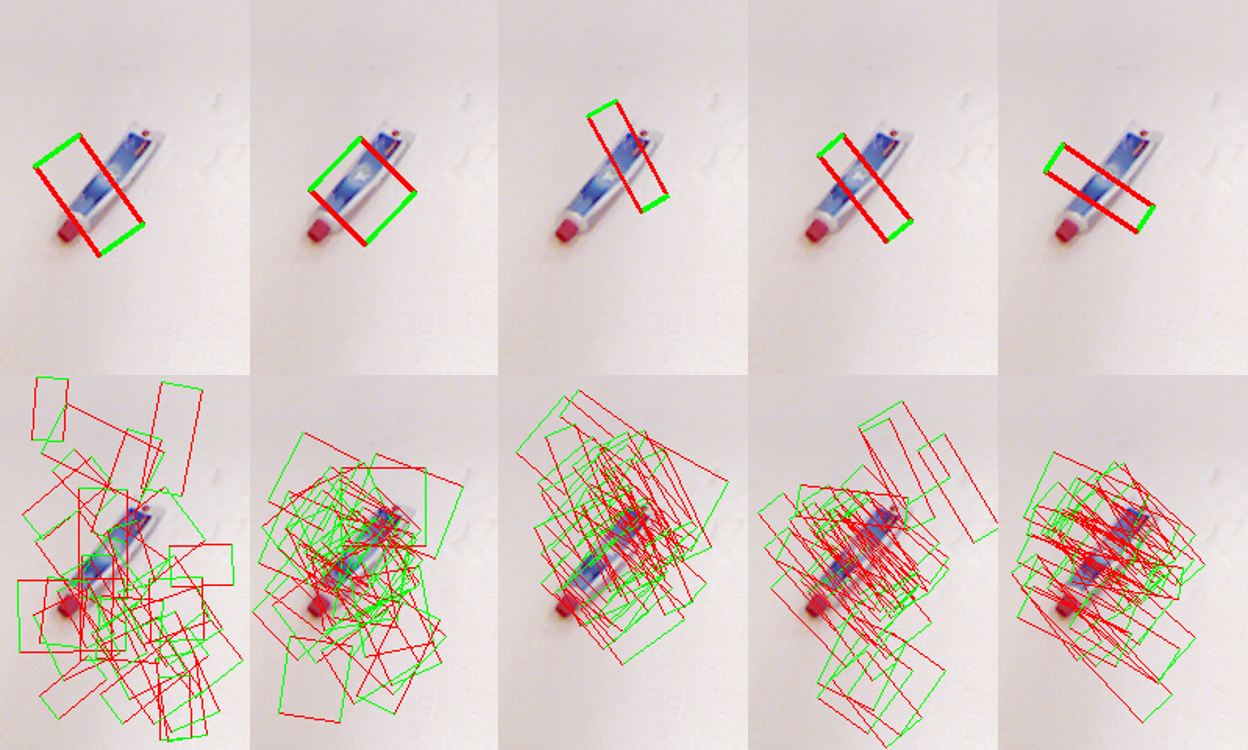}
	\caption{PSO iteration process.The first behavior of the image is the current global best position, and the second behavior is the current position of all particles.}
	\label{FIG:9}
\end{figure}

And if we make a small change to candidate estimator, the estimator can output the particles whose scores are above a certain threshold and rank high among all the particles at the same time so that we can multiple grasps.
\subsection{Improvements to PSO}
In order to make the PSO algorithm converge faster, we add some constraints to the particles during initialization or iteration. First, we hope that the particles are distributed as close to the object as possible. So when initializing particles, we distribute the particles within the center of the image and there has to be a particle with a score greater than 0.7. Second, we get the histogram of gray image corresponding to the RGB image before initializing the particles. According to the histogram which we get, we can roughly estimate the size of the target object. Then, we will initialize the particles differently according to the size of the target object. We can make sure the particle size is proper and reduce the effect of unproper particle size on particle score so that we can accelerate the convergence of the algorithm. Finally, we limit the size, aspect ratio and area of particles to a certain extent. If a particle is out of the limit during initialization or iteration, we will get the particle size back to a reasonable range by multiplying it by a corresponding correction factor.

\section{Experiment and evaluation}
\subsection{The Cornell Grasp Detection Dataset}
The dataset contains 885 images, 400 objects, 240 different categories and corresponding grasp labels. This data set is specially designed for parallel gripper. Each image contains multiple grasp labels. The labels are comprehensive and varied in orientation, location and scale, but it can't contain all possible grasps. And there are also errors in the labeling of partial positive samples and negative samples. Nonetheless, they are excellent examples of grasps. So we choose to analyze and evaluate based on the dataset.
\subsection{Image preprocessing}
Before inputting the image into the identification model, we preprocess the image. We clipped the original image to the center of the object, keeping 300 by 300 pixels. Then in order to compare with the previous work, we scaled the image to 224 by 224 pixels. 

As for the grasp identification model, we set the rectangular image horizontally along the long axis, fill the top and bottom ends with 0 pixels and scale the image to 24 by 24 pixels before inputting the feature image to the model.	Before training the model, we made necessary augmentation of the training set, including some operations such as translation, scaling, and rotation.
\subsection{The grasp identification model}
Before the input of the feature image into the model, we will preprocess the image, scale the image to 24*24 and fill it. So we can input the preprocessed feature image directly.

And our identification model outputs in the form of probability. And the output is probability of the network softmax layer classification of the graspable features (see Figure 10).
\begin{figure}[h]
	\centering
	\includegraphics[scale=.75,width=0.4\paperwidth]{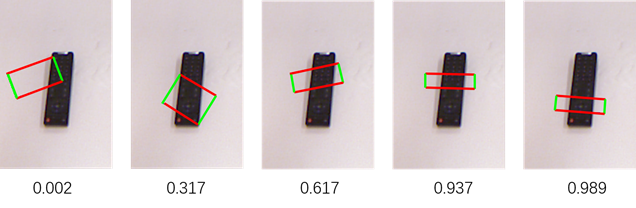}
	\caption{The output of the identification.}
	\label{FIG:10}
\end{figure}

If we use direct classification, take the third example in Figure 10. The model will classify it as graspable because the output is 0.617 which is greater than 0.5. If the model only outputs in the form of probability, it will not impact on subsequent iterations. So we don't directly evaluate models by how accurate they are, and it can be used as a reference for selecting identification model. 
\subsection{Multigrasp detection approach}
We have made a small modification to the original me\-thod. Instead of only printing the highest-scoring grasps, we output several of the highest-scoring grasps at the same time. These grasps converge to different positions of the target object, so we can have multiple ways to grab the same object.
\subsection{Grasp evaluation}
If we evaluate on the Cornell Grasp Detection Dataset, there have been two different evaluation criteria. The first approach is point evaluation, which evaluates whether the predicted grasp is a successful grasp through judging whether the distance between the center of the predicted fetch point and the center of the tag is below a certain threshold. There has been a lot of discussion about this evaluation criteria. The biggest problem with this evaluation criteria is that it doesn't take into account the size and angle of the grasp. And under this evaluation criteria, previous work has rarely revealed thresholds for evaluating comparisons. So it is difficult to compare results. Therefore, we do not use such evaluation criteria.
 
The second approach is rectangle evaluation. This approach considers the whole grasp rectangle. In this evaluation criteria, satisfying the following two points is considered to be a correct grab point:
\begin{enumerate}[(1)]
	\item The angle between the grasp and the graspable label shou\-ld not exceed
	\begin{math}
	30^{\circ}.
	\end{math}
	\item The intersection ratio of grasp and the graspable label should be no less than 20\%.
\end{enumerate}
In our work, we use the second approach to evaluate the model. Although we use a more proper approach to evaluate, there are still some problems. Although some predicted grasps are not evaluated to meet the above two points, they are actually graspable(see Figure 11).
\begin{figure}[h]
	\centering
	\includegraphics[scale=.75,width=0.4\paperwidth]{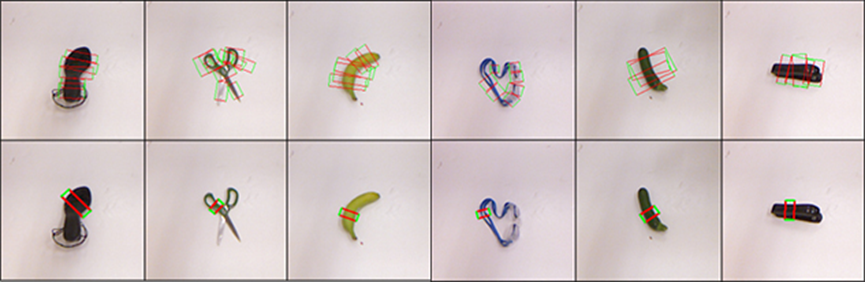}
	\caption{The first line is the grasps labeled by the dataset, and the second line is the result of our detection. Take the results of column 1 and column 4 from the left in the picture as an example. Although they do not meet the requirements of rectangle evaluation, it can still be considered to be graspable.}
	\label{FIG:11}
\end{figure}
\section{Results and comparisons}
We tested on a computer configured with a single CPU (i9-9900k 3.6G), a single GPU (NVDIA 1080Ti), and 16 gigabytes of memory. According to the evaluation criteria in part iii, we achieved good results in the Cornell dataset.
\subsection{Comparison of different approaches}
We compared the accuracy of our method with the results of previous experiments on the Cornell dataset (as shown in Table 1). Our approach achieved a success rate of 92.8\%. Among all known approaches, our success rate is only slightly lower than Guo et al. Compared to the existing two-stage method, our method improved the success rate by 11 percentage points.
\begin{table}[width=.8\linewidth,cols=2,pos=h]
\caption{We compare our approach with the previous work.}\label{tbl1}
\begin{tabular*}{\tblwidth}{@{} CCCC@{}}
\toprule
Algorithm & Accuracy(\%) \\
\midrule
Jiang et al.\cite{5980145} & 60.5\% \\
Lenz et al.\cite{1} & 73.9\% \\
Wang et al.\cite{doi:10.1177/1687814016668077} & 81.8\% \\
Redmon et al.\cite{2} & 88.0\% \\
Asif et al.\cite{7820198} & 88.2\% \\
Kumra et al.\cite{8202237} & 89.2\% \\
Guo et al.\cite{7989191} & 93.2\% \\
Our & 92.8\% \\
\bottomrule
\end{tabular*}
\end{table}

In terms of speed, our algorithm needs 378ms to process an image. Although our method is still much slower than the one-stage method of Kumra et al.\cite{8202237} and Redmon et al.\cite{2}, it should be able to run in real time.
\subsection{The result of multigrasp}
In the front, we mentioned that with a small change in our method, we can get multiple grab points at once, which can provide multiple solutions for grabbing an object. In this regard, we also carried out relevant experiments. The results are seen in Figure 12.
\begin{figure}
	\centering
	\includegraphics[scale=.75,width=0.38\paperwidth]{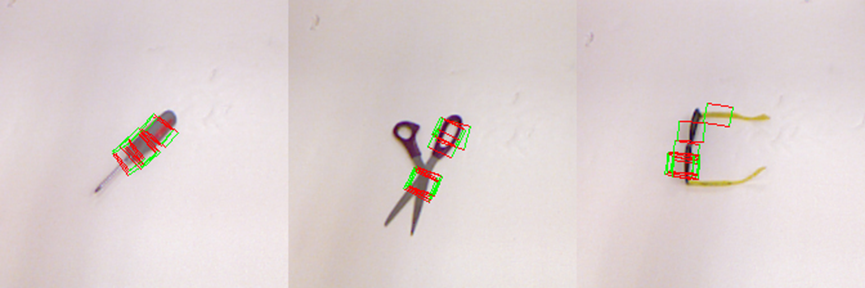}
	\caption{The result of the multigrasp approach.}
	\label{FIG:12}
\end{figure}
From the figure, we can see that the multigrasp method provides multiple correct grasps for each target object, and we also make statistics on the success rate and time consumption (see Table 2). In the evaluation of multigrasp detection, as long as one predicted grasp candidate of an image meets the requirements, we consider that the detection of capture points of this image is successful. It can be seen from the results that compared with the original method, the method of mutligrasp has a higher success rate. The increase in average time was only 5ms.
\begin{table}[width=.9\linewidth,cols=3,pos=h]
	\caption{We compare our approach with the previous work.}\label{tbl2}
	\begin{tabular*}{\tblwidth}{L@{} C@{} CC@{}}
		\toprule
		Algorithm & Accuracy(\%) &Average time(ms)\\
		\midrule 
		Original approach &92.8\% &378ms\\
		Multigrasp approach &94.8\% & 383ms\\		
		\bottomrule
	\end{tabular*}
\end{table}

In the multigrasp approach, we output multiple grasps at different locations. The fault tolerance of the identification model is further improved, because even if the model erroneously judges an ungraspable candidate to be the highest score of a target object, it is possible to find a correct grasp candidate from other grasps in the output.
\subsection{Summary}
We show that deep learning model can better learn grasping features, and PSO optimization algorithm can well solve the problem of multi-local optimal value. We used an uncomplicated network, greatly reducing computing costs and making the model easy to train and deploy. We turn the detection problem into an optimization problem for processing, so that the detection result is not too dependent on label data, but has multiple possibilities in the whole state space. We use PSO algorithm to deal with this kind of multi-variable complex function optimization problem with multi-local optimal advantages, which has low computational complexity, fast convergence speed and can find a better solution in considerable time. In the problem of grasp detection, we combined deep network learning characteristics and PSO optimization algorithm to solve the problem quickly, and obtained very advanced results.
\section{The experiment on the robot}
\subsection{The introduction of the experiment}
After our approach worked well on the lab computer, we decided to transplant it to a real robot for testing. The process is as follows (see Figure 13): first, we selected six objects that did not appear in the dataset. Then, we use the watershed algorithm to process the original image taken in the robot camera to extract the target object, and place the extracted object in the same position on the light white background image. Finally, we use our method to detect the capture point of the image obtained in the previous step, and draw the grasp on the original image.
\begin{figure}[h]
	\centering
	\includegraphics[scale=.75,width=0.4\paperwidth]{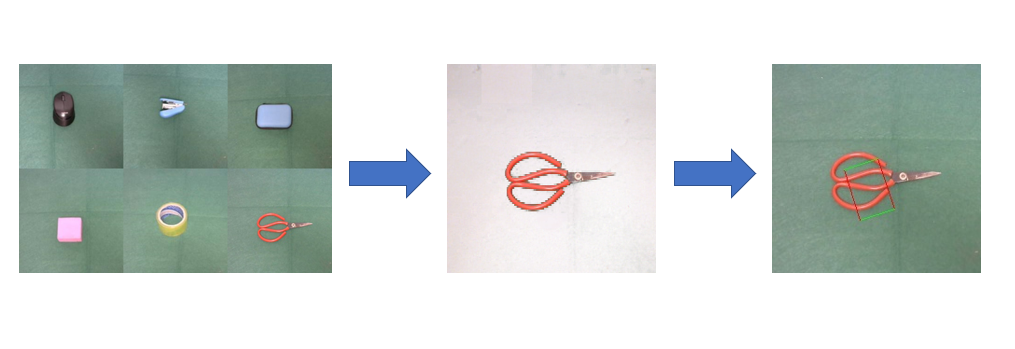}
	\caption{The process of experiment on the robot.}
	\label{FIG:13}
\end{figure}
\subsection{Result and discussion}
We used our method to detect the grasping points on the robot and obtained the following results (see Figure 14):
\begin{figure}[h]
	\centering
	\includegraphics[scale=.75,width=0.4\paperwidth]{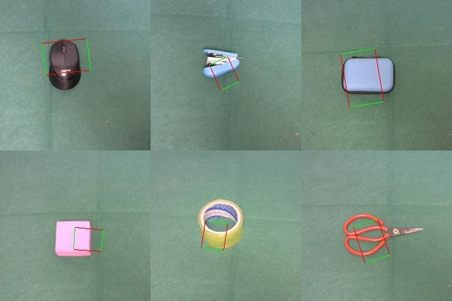}
	\caption{The result of the single grasp detection.}
	\label{FIG:14}
\end{figure}

While there are no major differences in kind between the objects we select and those in the Cornell dataset, there are significant differences in detail. However, this did not have a great impact on our detection results. Among the six selected objects, except the pink plastic blocks in the second row and the first column, the other predicted grasps were correct. This shows that our detection approach can not only be used on the robot, but also achieve good results in the case that the unfamiliar object is not marked, as long as the object in the data set belongs to the same category as the unfamiliar object. In practical use, we can select a few typical objects in the same kind of objects to mark accurately, so as to achieve a good effect in the detection of grasping points of most objects in this class, thus reducing the number of objects required by the data set, conducive to the application of our method in the actual detection.
\section{Conclusion}
In this article, we present a robotic grasp detection approach based on deep learning and PSO algorithm. We transfer the detection problem to an optimization problem. We first use the CNN to learn graspable feature. Then we take the identification model as the objective function of optimization, take the whole image as the state space, take the parameters of the rectangle identification points as variables and use the PSO algorithm to solve the optimization problem. In terms of accuracy, our algorithm is at the top of the Cornell Grasp Detection Dataset. Moreover, our approach is two-stage. Compared to end-to-end approach, it has a lower request to accuracy of the model and the dataset, which contributes to our practical use.

In the future work, we will improve the speed and robustness of the algorithm so that it can achieve better results in real industrial deployment. 
\appendix
\section{My Appendix}

All the codes used for experiment can be downloaded at \url{https://github.com/KathylinLawes/Robotic-grasp-detection}. And the color should be used for all figures in print.
\printcredits

\bibliographystyle{cas-model2-names}

\bibliography{cas-refs}

\newpage
\vskip1pt
\bio{}

\bio{figs/1.jpg}
Zhe Chu is an undergraduate at northwestern polytechnical university.
He came to northwestern polytechnical university in 2017. He is a member of the soccer robot base at northwestern polytechnical university.He is interested in robotics,machine learning and computer vision.
\endbio
\vspace{40pt} 
\bio{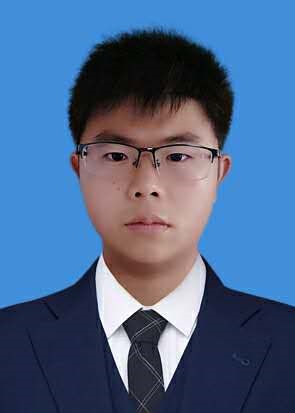}
Mengkai Hu received the bachelor’s degree in electronics and information engineering from Northwestern Polytechnical University in 2017. He is now pursuing the master's degree in School of Electronics Engineering and Computer Science, Peking University. His research focus on signal processing. 
\endbio

\vspace{30pt} 
\bio{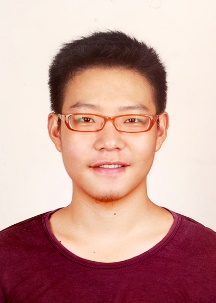}
Xiangyu Chen received the B.E. degree and the M.E. degree from Northwestern Polytechnical University, Xi’an, China, in 2017 and 2020 respectively. Currently he is a research assistant in Shenzhen Institutes of Advanced Technology, Chinese Academy of Science. His research focus on robotics, deep learning and computer vision.
\endbio

\end{document}